\documentclass[final]{cvpr}

\usepackage{times}
\usepackage{epsfig}
\usepackage{graphicx}
\usepackage{amsmath}
\usepackage{amssymb}

\usepackage[pagebackref=true,breaklinks=true,colorlinks,bookmarks=false]{hyperref}

\usepackage[utf8]{inputenc} 
\usepackage[T1]{fontenc}  
\usepackage{hyperref}    
\usepackage{url}      
\usepackage{booktabs}    
\usepackage{amsfonts}    
\usepackage{nicefrac}    
\usepackage{microtype}   

\usepackage{xspace}
\usepackage[export]{adjustbox}
\usepackage{color}
\usepackage{lmodern}
\usepackage{xcolor}
\usepackage{cuted}
\usepackage{multirow}

\newcommand{\dsname}[1]{\texttt{\small #1}\xspace}
\newcommand{\dsnamebold}[1]{{\fontseries{b}\selectfont\texttt{#1}}\xspace}
\newcommand{\funcA}{\dsname{ACON-A}}
\newcommand{\funcB}{\dsname{ACON-B}}
\newcommand{\funcC}{\dsname{ACON-C}}
\newcommand{\func}{\dsname{ACON}}
\newcommand{\funcX}{\dsname{meta-ACON}}
\newcommand{\mbR}{\mathbb{R}}
\newcommand{\funcF}{\dsname{ACON-FReLU}}

\newcommand{\funcbA}{\dsnamebold{ACON-A}}
\newcommand{\funcbB}{\dsnamebold{ACON-B}}
\newcommand{\funcbC}{\dsnamebold{ACON-C}}
\newcommand{\funcb}{\dsnamebold{ACON}}
\newcommand{\funcbX}{\dsnamebold{Meta-ACON}}

\newcommand{\fnet}{TFNet}

\usepackage{array}
\newcommand{\PreserveBackslash}[1]{\let\temp=\\#1\let\\=\temp}
\newcolumntype{C}[1]{>{\PreserveBackslash\centering}p{#1}}
\newcolumntype{R}[1]{>{\PreserveBackslash\raggedleft}p{#1}}
\newcolumntype{L}[1]{>{\PreserveBackslash\raggedright}p{#1}}

\def\cvprPaperID{1454} 
\def\confYear{CVPR 2021}

\begin{document}

\title{Activate or Not: Learning Customized Activation }

	\author{Ningning Ma\textsuperscript{\rm 1}\qquad~Xiangyu~Zhang\textsuperscript{\rm 2}\qquad~Ming~Liu\textsuperscript{\rm 1}\qquad~Jian~Sun\textsuperscript{\rm 2}\\
		{\textsuperscript{\rm 1} 
The Hong Kong University of Science and Technology}  \\ 
		{\textsuperscript{\rm 2}MEGVII Technology} \\
		\small{\texttt{\{nmaac,eelium\}@ust.hk}} \qquad
		\small{\texttt{\{zhangxiangyu,sunjian\}@megvii.com}} 
	}

\maketitle

\footnotetext[1]{This work is supported by The National Key Research and Development Program of China (No. 2017YFA0700800) and Beijing Academy of Artificial Intelligence (BAAI).}

\begin{abstract}
 We present a simple, effective, and general activation function we term \func which learns to activate the neurons or not.
 Interestingly, we find Swish, the recent popular NAS-searched activation, can be interpreted as a smooth approximation to ReLU.
 Intuitively, in the same way, we approximate the more general Maxout family to our novel \func family, which  remarkably improves the performance and makes Swish a special case of \func.
  Next, we present \funcX, which \textbf{explicitly} learns to optimize the parameter switching between non-linear (activate) and linear (inactivate) and provides a new design space.
 By simply changing the activation function, we show its effectiveness on both small models and highly optimized large models (e.g. it improves the ImageNet top-1 accuracy rate by 6.7\% and 1.8\% on MobileNet-0.25 and ResNet-152, respectively). Moreover, our novel \func can be naturally transferred to object detection and semantic segmentation, showing that \func is an effective alternative in a variety of tasks. Code is available at \url{https://github.com/nmaac/acon}. 
\end{abstract}


\section{Introduction}
The Rectified Linear Unit (ReLU) \cite{hahnloser2000digital,jarrett2009best,nair2010rectified} has become an effective component in neural networks and a foundation of many state-of-the-art computer vision algorithms.
Through a sequence of advances, the Swish activation \cite{ramachandran2017searching} searched by the Neural Architecture Search (NAS) technique achieves top accuracy on the challenging ImageNet benchmark \cite{deng2009imagenet,russakovsky2015imagenet}.
It has been shown by many practices to ease optimization and achieve better performance \cite{howard2019searching,tan2019efficientnet}.
Our goal is to interpret the mechanism behind this searched result and investigate more effective activation functions.

\begin{figure}
 \centering
 \includegraphics[width=0.8\linewidth]{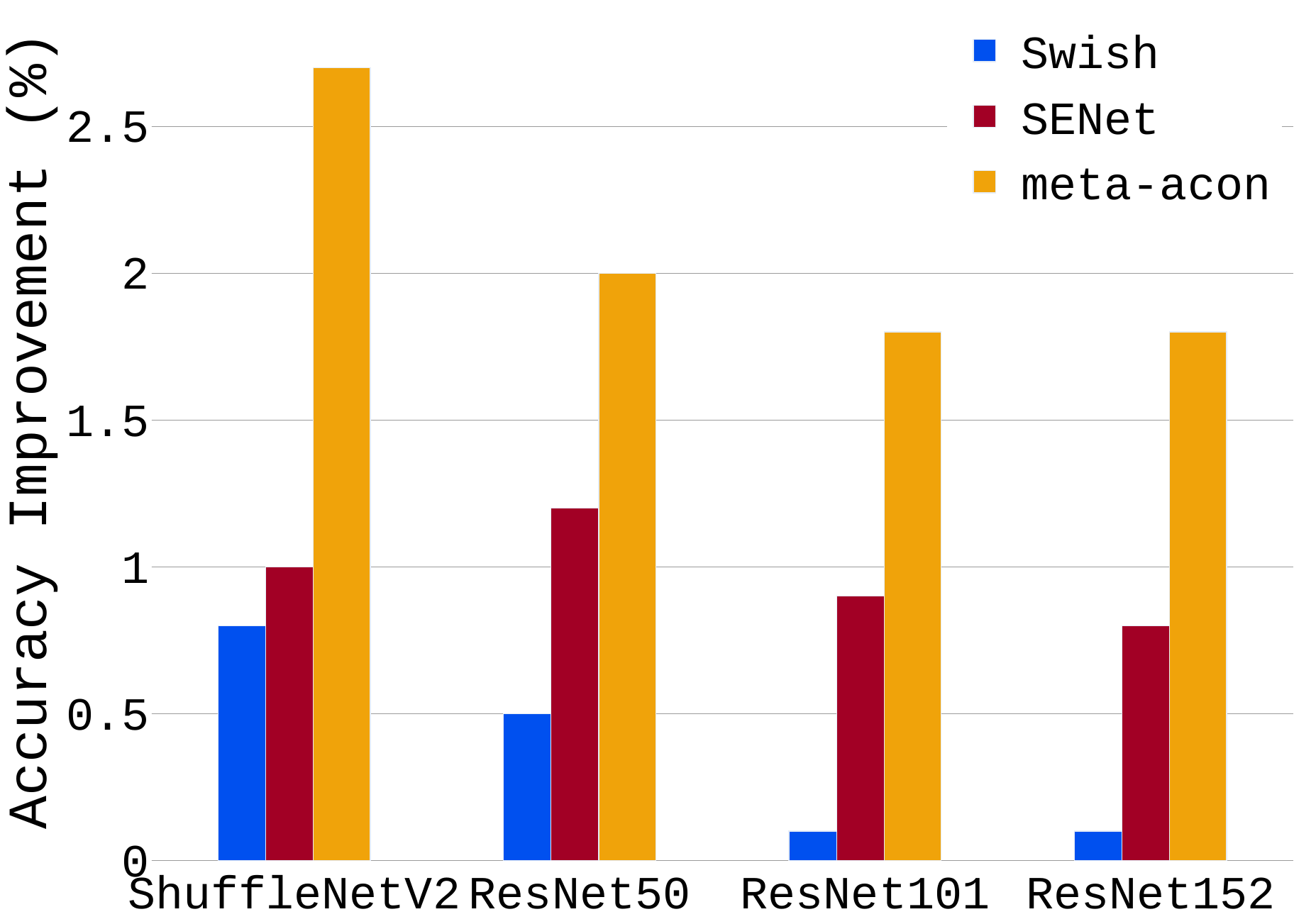}
 \caption{ImageNet top-1 accuracy relative improvements compared with the ReLU baselines. As the models go larger, Swish and SENet gain smaller, but \funcbX still improves very stably and remarkably even on the substantially deep and highly optimized ResNet152.
 The relative improvements of \funcbX are about twice as much as SENet. }
 \label{fig:result}
\end{figure}

\begin{figure*}[t]
 \centering
 \includegraphics[width=0.8\linewidth]{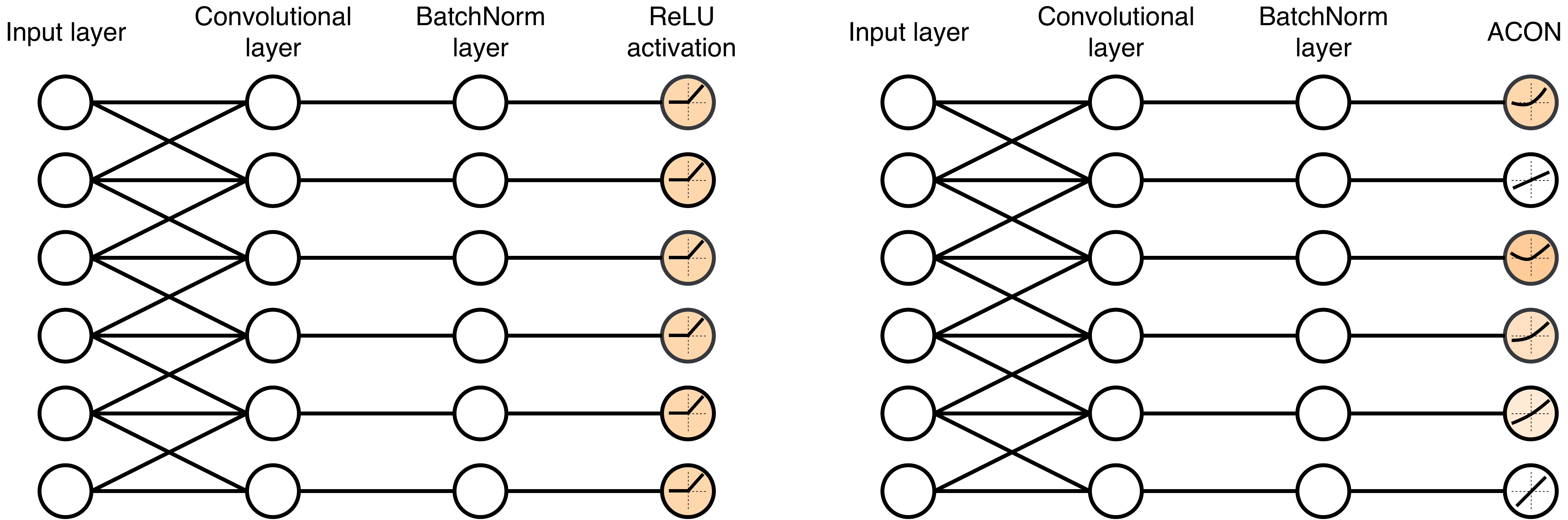}
 \caption{We propose a novel activation function we term the \func that $explicitly$ learns to activate the neurons or not. \textbf{Left:} A ReLU network; \textbf{Right:} An \func network that learns to activate (orange) or not (white). }
 \label{fig:acon}
\end{figure*}

Despite the success of NAS on modern activations, a natural question to ask is: \textit{how does the NAS-searched Swish actually work?} Despite its widespread use, this activation function is still poorly understood. 
We show that Swish can be surprisingly represented as a smooth approximation to ReLU, by a simple and general \textit{approximation formula} (Equ. \ref{equ:trans}).

\begin{figure*}[t]
 \centering
 \includegraphics[width=0.85\linewidth]{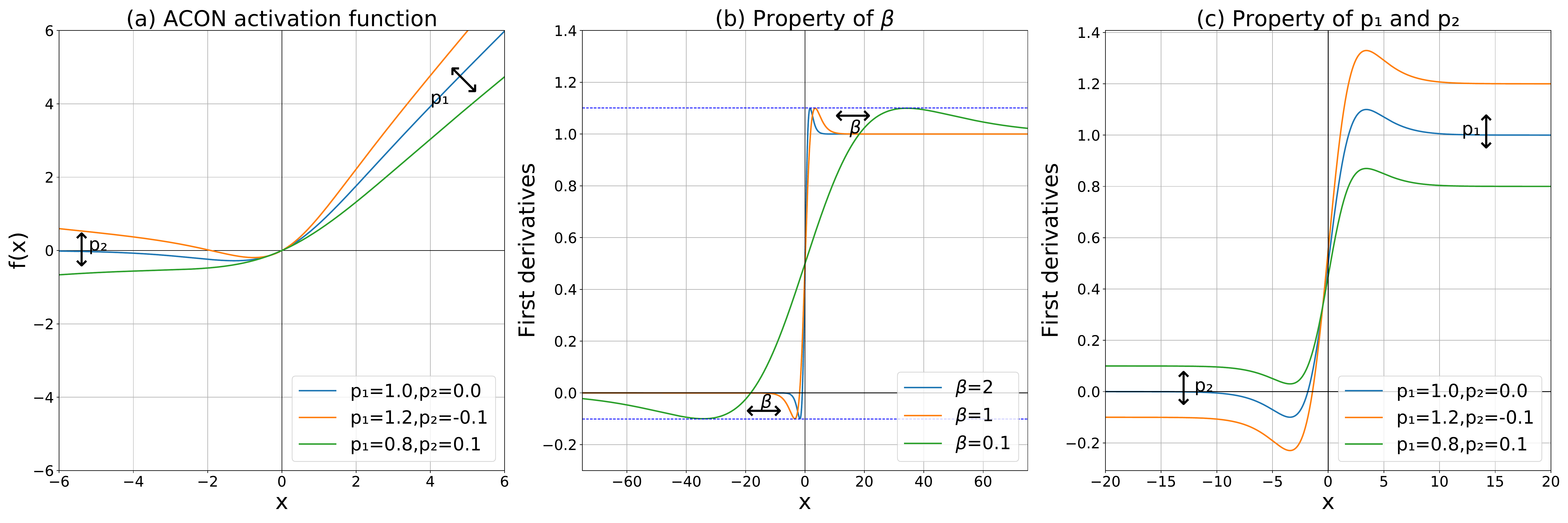}
 \caption{The \func activation function and its first derivatives. (a) The \funcC activation function with fixed $\beta$ (see Fig. \ref{fig:smoothmax} for the influence of $\beta$);
  (b-c) The first derivatives with fixed $p_1 \& p_2$ (b) and fixed $\beta$ (c). $\beta$ controls how fast the first derivative asymptotes to the upper/lower bounds, which are determined by $p_1$ and $p_2$.}
 \label{fig:curve}
\end{figure*}

This paper pushes the envelop further: our method, called \func, follows the spirit of the ReLU-Swish conversion and approximates the general Maxout \cite{goodfellow2013maxout} family to our novel \func family by the general \textit{approximation formula}.
We show the converted functions (\funcA, \funcB, \funcC) are smooth and differentiable, where Swish is merely a case of them (\funcA).
\func is conceptually simple and does not add any computational overhead, however, it improves accuracy remarkably.
To achieve this result, we identify the fixed upper/lower bounds in the gradient as the main obstacle impeding from improving accuracy and present the \func with learnable upper/lower bounds (see Fig. \ref{fig:curve}). 

In principle, \func is an extension of Swish and has a dynamic non-linear degree, where a switching factor decays to zero as the non-linear function becomes linear. 
Intuitively, this switching factor enables \func to switch between activating or not.
However, evidence \cite{ramachandran2017searching} has shown that optimizing the factor simply by using the gradient descent cannot learn to switch between linear and non-linear well.
Therefore, we optimize the switching factor $explicitly$ for fast learning and present \funcX that learns to learn whether to activate or not (see Fig. \ref{fig:acon}).
Despite it seems a minor change, \funcX has a large impact: it has significant improvements on various tasks very stably (even the highly-optimized and extremely deep SENet-154) and provides a new architecture design space in the meta learner, which could be layer-wise, channel-wise, or pixel-wise. The design in the provided space is beyond the focus of this paper, but it is suggestive for future research.

\func transfers well on a wide range of tasks.
No matter for small models or large models, our approach surpasses the ReLU counterpart significantly: it improves the ImageNet top-1 accuracy rate by 6.7\% and 1.8\% on MobileNet-0.25 and ResNet-152, respectively.
We show its generality on object detection and semantic segmentation tasks.

We summarize our contributions as follows:
(1) We present a novel perspective to understand Swish as a smoothed ReLU;
(2) from this valuable perspective, we connect the two seemingly unrelated forms (ReLU and Swish), and smooth ReLU’s general Maxout family to Swish’s general \func family; 
(3) we present \funcX that explicitly learns to activate the neurons or not, improves the performance remarkably.

\section{Related Work}
\paragraph{Activation functions}

The Rectified Linear Unit (ReLU) \cite{hahnloser2000digital,jarrett2009best,nair2010rectified} and its variants \cite{maas2013rectifier,he2015delving,clevert2015fast,ma2020funnel} are the most widely used activation functions in the past few years.
ReLU is non-differentiable at zero and is differentiable anywhere else. 
Many advances followed \cite{klambauer2017self,singh2019filter,agostinelli2014learning,hendrycks2016bridging,qiu2018frelu,elfwing2018sigmoid,xu2015empirical}, 
and softplus \cite{Dugas2000IncorporatingSF} is a smooth approximation to the maximum function ReLU based on the LogSumExp function.

The Maxout \cite{goodfellow2013maxout} can be a piecewise linear approximation for arbitrary convex activation functions. It generalizes the leaky ReLU and ReLU and can approximate linear activations.
The Maxout is a general formulation of many current activation functions. In this work we present a new family of activation which is a smooth approximation to the Maxout family. 
For example, the recent searching technique contributes to a new searched scalar activation called Swish \cite{ramachandran2017searching} by combing a comprehensive set of unary functions and binary functions.
Firstly, the searched results show that the form of SiLU \cite{elfwing2018sigmoid,hendrycks2016bridging}, $y=x \cdot Sigmoid(x)$, achieves good performance and outperforms other scalar activations on many vision tasks.
Secondly, the method also indicates that the form of Swish, $y=x \cdot Sigmoid(\beta x)$, shows great potential.
However, there is a lack of proper explanation of the searched Swish formulation. 
In this paper, we generalize Swish to the \func family, showing that it is a smooth approximation of ReLU based on the well-known smooth conversion called $\alpha$-$softmax$, which is frequently applied in optimization and neural computation \cite{lange2014applications,Boyd2006ConvexO,Haykin1998NeuralNA}.

A recent activation called DY-ReLU \cite{chen2020dynamic} encodes the global context into a hyperfunction and adapts the piecewise linear activation function accordingly.
The method increases the number of parameters remarkably and improves the performance significantly especially on light-weight models. However, the improvements become smaller when the models go larger and deeper. For example, though DY-ReLU generalizes SENet conceptually, ResNet50-DY-ReLU (only 1.0\% improvement, 76.2->77.2) cannot outperform ResNet50-SENet, the improvements on larger models become smaller.
Different from DY-ReLU, firstly, \func learns to determine the activation to be linear or non-linear.
Secondly, our method has a comparable amount of parameters with ReLU networks. Thirdly, the performance improvement is still very significant even on the very deep and highly optimized ResNet-152, which is more than twice as much as SENet ($\Delta$ accuracy = 1.8 v.s. 0.8, see Fig. \ref{fig:result}).

\paragraph{Dynamic network}
Standard CNNs \cite{simonyan2014very,he2016deep,szegedy2015going,zhang2018shufflenet,chollet2017xception,howard2017mobilenets,sandler2018mobilenetv2,howard2019searching} share the same network structure and convolution kernels for all the samples, while conditional (or dynamic) CNNs \cite{lin2017runtime,liu2018dynamic,wu2018blockdrop,yu2018slimmable,keskin2018splinenets,ma2020weightnet} use dynamic kernels, widths, or depths conditioned on the input samples, obtaining remarkably gains in accuracy.

Some dynamic networks learn the dynamic kernels \cite{yang2019condconv,ma2020weightnet}, some use attention-based approaches \cite{vaswani2017attention,luong2015effective,bahdanau2014neural,wang2017residual,woo2018cbam,hu2018squeeze} to change the network structures, another series of work \cite{wang2018skipnet,huang2017multi} focus on dynamic depths of the convolutional networks, that skip some layers for different samples.
In our work, we learn the non-linear degree in the activation function dynamically, which controls to what degree the non-linear layer is.

\paragraph{Neural network design space}
The design of the neural network architecture mainly includes the kernel level space and the feature level space.
The most common feature design space aims to optimize the performance via channel dimension \cite{szegedy2015going,szegedy2016rethinking,sandler2018mobilenetv2,ma2018shufflenet,hu2018squeeze}, spatial dimension \cite{tan2019efficientnet,cao2019gcnet,jaderberg2015spatial} and feature combination \cite{he2016deep,Huang2016DenselyCC}.
On the recent popular kernel design space we can optimize the kernel shape \cite{szegedy2017inception,yu2015multi,jeon2017active} and kernel computation \cite{yang2019condconv,ma2020weightnet}. 
In this work, we provide a new design space on the non-linear degree level by customizing the non-linear degree in each layer.

\section{ACON}

\begin{table*}[h] 
\caption{Summary of the Maxout family and \func family. $\sigma$ denotes Sigmoid.}
 \label{table:static}
 \centering
\begin{tabular}{ll|l|l}
  \toprule
    &     & Maxout family             & \func family                     \\ \midrule
$\eta_a(x)$   & $\eta_b(x)$   & $max(\eta_a(x),\eta_b(x))$          & $(\eta_a(x)-\eta_b(x))\cdot \sigma(\beta (\eta_a(x)-\eta_b(x)))+\eta_b(x)$    \\ \midrule
$x$   & $0$   & $max(x,0):ReLU$    & $\funcA (Swish): x \cdot \sigma(\beta x)$       \\
$x$   & $px$  & $max(x, px):PReLU$ & $\funcB: (1-p)x \cdot \sigma(\beta (1-p)x)+px$ \\
$p_1x$ & $p_2x$ & $max(p_1x, p_2x)$     &  $\funcC: (p_1-p_2)x \cdot \sigma( \beta (p_1-p_2)x)+p_2x$       \\
  \bottomrule
\end{tabular}
\end{table*}

We present \textbf{Ac}tivate\textbf{O}r\textbf{N}ot (\func) as a way of learning to activate the neurons or not.
In this paper, we first show how we use the general approximation formula: \textit{smooth maximum} \cite{lange2014applications,Boyd2006ConvexO,Haykin1998NeuralNA} to perform the ReLU-Swish \cite{ramachandran2017searching} conversion.
Next, we convert other cases in the general Maxout \cite{goodfellow2013maxout} family, which is thus a natural and intuitive idea and makes Swish a case of \func.
At last, \func learns to activate (non-linear) or not (linear) by simply maintaining a switching factor, we introduce \funcX that learns to optimize the factor explicitly and shows significant improvements.

\paragraph{Smooth maximum}
We begin by briefly reviewing the smooth maximum function.
Consider a standard maximum function $max(x_1,...,x_n)$ of $n$ values, we have its smooth and differentiable approximation:
\begin{equation}
S_{\beta}(x_1,...,x_n)=\frac{\sum_{i=1}^{n}{x_i e^{\beta x_i}}}{\sum_{i=1}^{n}{e^{\beta x_i}}}
\end{equation}

where $\beta$ is the switching factor: when $\beta \rightarrow \infty$, $S_\beta \rightarrow max$; when $\beta \rightarrow 0$, $S_\beta \rightarrow arithmetic\ mean$.

In neural networks, many common activation functions are in the form of $max(\eta_a(x),\eta_b(x))$ function (e.g. ReLU $max(x,0)$ and its variants) where $\eta_a(x)$ and $\eta_b(x)$ denote linear functions.
Our goal is to approximate the activation functions by this formula.
Therefore we consider the case when $n=2$, we denote $\sigma$ as the Sigmoid function and the approximation becomes:
\begin{equation}
\label{equ:trans}
\begin{split}
&S_{\beta}(\eta_a(x),\eta_b(x)) \\
&=\eta_a(x)\cdot \frac{e^ {\beta \eta_a(x)}}{e^{\beta \eta_a(x)}+e^{\beta \eta_b(x)}} + \eta_b(x) \cdot \frac{e^ {\beta \eta_b(x)}}{e^{\beta \eta_a(x)}+e^{\beta \eta_b(x)}} \\
&=\eta_a(x)\cdot \frac{1}{1+e^{-\beta (\eta_a(x)-\eta_b(x))}}+\eta_b(x)\cdot \frac{1}{1+e^{-\beta (\eta_b(x)-\eta_a(x))}} \\  
&=\eta_a(x)\cdot \sigma [\beta(\eta_a(x)-\eta_b(x))]+\eta_b(x) \cdot \sigma [\beta(\eta_b(x)-\eta_a(x))] \\
&=(\eta_a(x)-\eta_b(x))\cdot \sigma [\beta(\eta_a(x)-\eta_b(x))]+\eta_b(x)
\end{split}
\end{equation}

\paragraph{\funcbA} 
We consider the case of ReLU when $\eta_a(x)=x, \eta_b(x)=0$, then $ f_\funcA(x) = S_\beta(x,0)=x\cdot \sigma(\beta x)$, 
which we call \funcA and is exactly the formulation of Swish \cite{ramachandran2017searching}. 
Swish is a recent new activation which is a NAS-searched result, although it is widely used recently, there is a lack of reasonable explanations about why it improves the performance.
From the perspective above, we observe that Swish is a smooth approximation to ReLU.

\paragraph{\funcbB} 
Intuitively, based on the approximation we could convert other maximum-based activations in the Maxout family (e.g. Leaky ReLU \cite{maas2013rectifier}, PReLU \cite{he2015delving}, etc) to the \func family.
Next, we show the approximation of PReLU.
It has an original form $f(x)=max(x,0)+p\cdot min(x,0)$, where $p$ is a learnable parameter and initialized as 0.25. However, in most case $p<1$, under this assumption, we rewrite it to the form: $f(x)=max(x, px) (p<1)$.
Therefore we consider the case when $\eta_a(x)=x, \eta_b(x)=px$ in Equ. \ref{equ:trans} and get the following new activation we call \funcB:
\begin{equation}
f_\funcB(x) = S_\beta(x,px)=(1-p)x \cdot \sigma[ \beta(1-p)x]+px
\end{equation}

\paragraph{\funcbC} 
Intuitively, we present a simple and more general case we term \funcC.
We adopt the same two-argument function, with an additional hyper-parameter.
\funcC follows the spirit of \funcB that simply uses hyper-parameters scaling on the feature. Formally, let $\eta_a(x)=p_1x, \eta_b(x)=p_2x (p_1\ne p_2)$:
\begin{equation}
f_\funcC(x) = S_\beta(p_1x,p_2x)=(p_1-p_2)x \cdot \sigma[\beta(p_1-p_2)x]+p_2x
\end{equation}

\begin{figure}
\centering
\includegraphics[width=.36\textwidth,center]{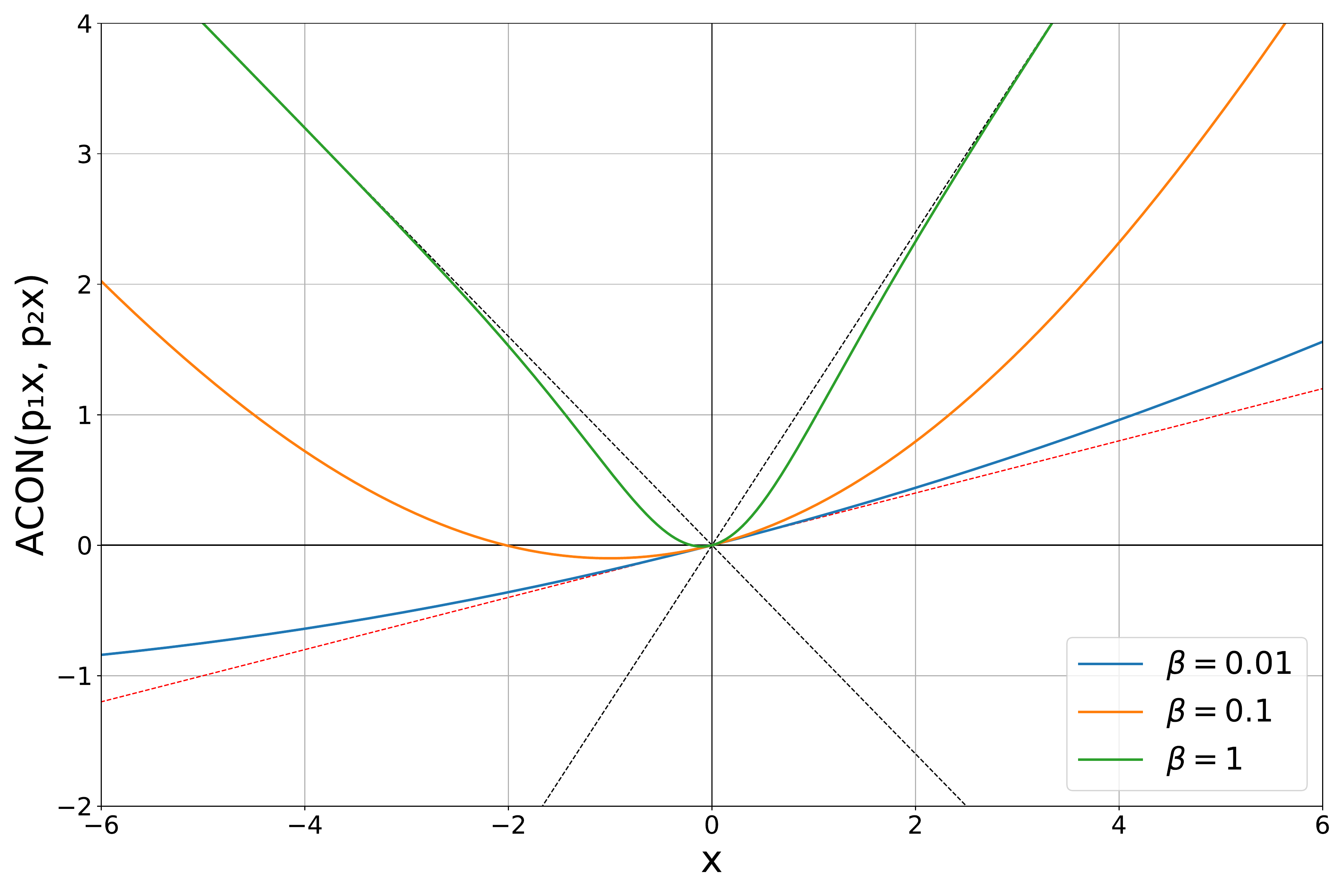}
 \caption{\funcC applied on '$1.2x$' and '$-0.8x$' functions with various values of $\beta$, which switches between non-linear and linear (arithmetic mean). 
 Very large $\beta$ can approximate the $maximum$ function (non-linear) and $\beta$ near zero can approximate the $mean$ function (linear) (the red dashed line).
 }
 \label{fig:smoothmax}
\end{figure}

As with PReLU, $\beta, p_1$, and $p_2$ are channel-wise. We init $\beta$=$p_1$=1, $p_2$=0.
Our definition of \funcC is a very simple and general case (see Fig. \ref{fig:curve}, Fig. \ref{fig:smoothmax}).
Moreover, there could be many more complicated cases in the Maxout family (e.g. more complicated formulations of $\eta_a(x)$ and $\eta_b(x)$), which are beyond the scope of this paper.
We focus on the property of the conversion on this simple form.

\paragraph{Upper/lower bounds in the first derivative.} 
We show that Swish has fixed upper/lower bounds (Fig. \ref{fig:curve} b) but our definition of \funcC allows the gradient has learnable upper/lower bounds (Fig. \ref{fig:curve} c).
Formally, we compute the first derivative of \funcC and its limits as follows:
\begin{equation}
\begin{split}
\label{equ:first}
&\frac{d }{dx} [f_\funcC(x)]\\ 
=& \frac{ (p_1-p_2)(1+e^{-\beta (p_1x-p_2x)})} {(1+e^{-\beta (p_1x-p_2x)})^2} \\
+& \frac{\beta (p_1-p_2)^2 e^{-\beta (p_1x-p_2x))} x} {(1+e^{-\beta (p_1x-p_2x)})^2}+p_2 \\
\end{split}
\end{equation}

\begin{equation}
\lim_{x\rightarrow \infty} \frac{d f_\funcC(x)}{dx} = p_1, \ \ \lim_{x\rightarrow -\infty} \frac{d f_\funcC(x)}{dx} = p_2 \ \ (\beta>0)
\end{equation}

To compute the upper/lower bounds, which are the maxma/minima values, we compute the second derivative:
\begin{equation}
\begin{split}
&\frac{d^2 }{dx^2} [f_\funcC(x)] \\
=& \beta \left(p_2-p_1\right)^2\mathrm{e}^{\beta \left(p_1-p_2\right)x} \cdot \\
&\dfrac{\left(\left(\beta \left( p_2- p_1\right)x+2\right)\mathrm{e}^{\beta \left(p_1-p_2\right)x}+\beta \left( p_1- p_2\right)x+2\right)}{\left(\mathrm{e}^{\beta \left(p_1-p_2\right)x}+1\right)^3} 
\end{split}
\end{equation}
We set $\frac{d^2 }{dx^2} [f_\funcC(x)]=0$, simplify it, and get $(y-2)e^y=y+2$, where $y=(p_1-p_2)\beta x$.
Solving the equation we get $y\approx \pm 2.39936$. Then we get maxima and minima of Equ. \ref{equ:first} when $\beta>0$: 
\begin{equation}
\begin{split}
maxima( \frac{d }{dx} [f_\funcC(x)] ) &\approx 1.0998 p_1 - 0.0998 p_2, \\
minima( \frac{d }{dx} [f_\funcC(x)]  ) &\approx 1.0998 p_2 - 0.0998 p_1 \ \  \\
\end{split}
\end{equation}
This is different from Swish with the fixed upper/lower bounds (1.0998, -0.0998) in the first derivative.
In Swish, the hyper-parameter $\beta$ only determines how fast the first derivative asymptotes to the upper bound and the lower bound, however, the bounds are learnable and determined by $p_1$ and $p_2$ in \funcC (see Fig. \ref{fig:curve} c).
The learnable boundaries are essential to ease optimization and we show by experiments that these learnable upper/lower bounds are the key for improved results.

\subsection{Meta-ACON}
\label{section:meta}

\func switches the activation to activate or not as the switching factor $\beta$ controls it to be non-linear or linear.
Specifically, when $\beta \rightarrow \infty, f_\funcC(x) \rightarrow max(p_1x,p_2x)$; when $\beta \rightarrow 0, f_\funcC(x) \rightarrow mean(p_1x,p_2x)$.
Thus, unlike the traditional activations such as the ReLU, \func allows each neuron to adaptively activate or not (see Fig. \ref{fig:acon}).  
This customized activating behavior helps to improve generalization and transfer performance.
This motivated us to develop the following \funcX that plays a key role in the customized activation.

Our proposed concept is simple: learning the switching factor $\beta$ explicitly conditioned on the input sample $x \in \mbR^{C \times H \times W}$: $\beta = G(x)$.
We are not aiming to propose a specific structure, we provide a design space in the generating function $G(x)$.

\paragraph{Design space}
The concept is more important than the specific architecture which can be layer-wise, channel-wise of pixel-wise structure. 
Our goal is to present some simple designing examples, which manage to obtain significantly improved accuracy and show the importance of this new design space.

We briefly use a routing function to compute $\beta$ conditioned on input features and present some simple structures.
First, the structure can be layer-wise, which means the elements in a layer share the same switching factor. Formally, we have:
$\beta=\sigma \sum_{c=1}^{C}\sum_{h=1}^{H}\sum_{w=1}^{W}{x_{c,h,w}}$.

Second, we present a simple channel-wise structure, meaning the elements in a channel share the same switching factor. Formally, we show it by:
$\beta_c=\sigma W_1 W_2 \sum_{h=1}^{H}\sum_{w=1}^{W}{x_{c,h,w}}$.
We use $W_1 \in \mbR^{C \times C/r}$, $W_2 \in \mbR^{C/r \times C}$ to save parameters ($r=16$ by default).
To further reduce the amount of parameters for large models (Res152), we find the depth-wise fully-connected layers also achieve good results.

Third, for the pixel-wise structure, all the elements use unique factors. 
Although there could be many structure designing methods, we simply present an extremely simple structure aiming to present a pixel-wise example.
Formally, we have:
$\beta_{c,h,w}=\sigma {x_{c,h,w}}$.

We note our \funcX has a straightforward structure. 
For the following \funcX experiments, we use the channel-wise structure and \funcC unless otherwise noted.
More complex designs have the potential to improve performance but are not the focus of this work.

\begin{figure}
\centering
\includegraphics[width=.38\textwidth,center]{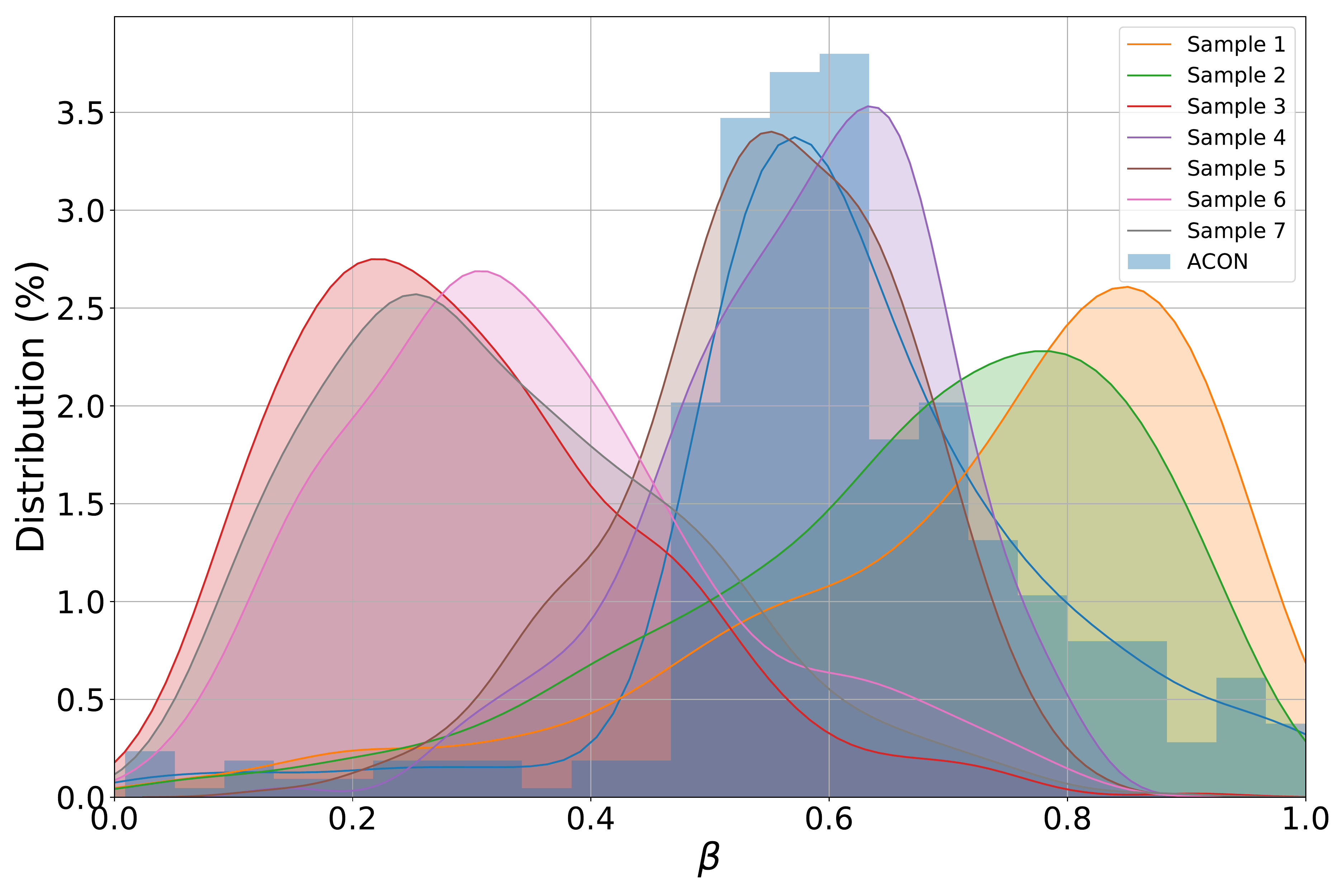}
 \caption{\func v.s. \funcX. $\beta$-distribution of an activation layer in the last bottleneck of the trained ResNet-50 models. 
 We randomly select 7 samples, for the \func network, they share the same $\beta$ distribution (the blue histogram); while for the \funcX network, they have 7 distinct $\beta$ distributions. 
  Lower $\beta$ values mean more linear, lager $\beta$ values mean more non-linear.}
 \label{fig:distribution}
\end{figure}

\section{Experiment}

\subsection{Image Classification}
We present a thorough experimental comparison on the challenging ImageNet 2012 classification dataset \cite{deng2009imagenet,russakovsky2015imagenet} along with comprehensive ablations.
For training, we follow the common practice and train all the models using the same input size of 224x224 and report the standard top-1 error rate.

\begin{table*}
 \caption{\textbf{Comparison of the \funcX on MobileNets, ShuffleNetV2, and ResNets.} We report the top-1 error on the ImageNet dataset (train and test on 224x224 input size).}
 \label{table:dynamic}
 \centering
 \begin{tabular}{l|c|c|c|c|c|c}
  \toprule
  & \multicolumn{3}{c|}{ReLU} & \multicolumn{3}{c}{\funcX} \\ \midrule
 & FLOPs & \# Params. & Top-1 err. & FLOPs & \# Params. & Top-1 err. \\ \midrule
MobileNetV1 0.25    & 41M   & 0.5M  &47.6  & 41M & 0.6M   &  $\mathbf{40.9_{(+6.7)}}$      \\ 
MobileNetV2 0.17    & 42M   & 1.4M  &52.6  & 42M & 1.9M   &  $\mathbf{46.2_{(+6.4)}}$      \\ 
ShuffleNetV2 0.5x    & 41M   & 1.4M  &39.4  & 41M & 1.7M   &  $\mathbf{34.8_{(+4.6)}}$      \\ 
MobileNetV1 0.75    & 325M  & 2.6M  &30.2  & 326M & 3.1M   & $\mathbf{26.4_{(+3.8)}}$      \\ 
MobileNetV2 1.0    & 299M  & 3.5M  &27.9  & 299M & 6.0M   &  $\mathbf{25.0_{(+2.9)}}$      \\ 
ShuffleNetV2 1.5x    & 299M  & 3.5M  &27.4  & 299M & 3.9M   &  $\mathbf{24.7_{(+2.7)}}$      \\ \midrule
ResNet-18    & 1.8G  & 11.7M  &30.3  & 1.8G & 11.9M   &  $\mathbf{28.4_{(+1.9)}}$      \\ 
ResNet-50    & 3.9G  & 25.5M  &24.0  & 3.9G & 25.7M   &  $\mathbf{22.0_{(+2.0)}}$      \\ 
ResNet-101    & 7.6G  & 44.4M  &22.8  & 7.6G & 44.8M   &  $\mathbf{21.0_{(+1.8)}}$      \\ 
ResNet-152    & 11.3G  & 60.0M  &22.3  & 11.3G & 60.5M   &  $\mathbf{20.5_{(+1.8)}}$      \\ 

  \bottomrule
 \end{tabular}
\end{table*}

\begin{table*}
 \caption{\textbf{Comprehensive comparison of \func on ShuffleNetV2 and ResNets.} We report the top-1 error on the ImageNet dataset (train and test on 224x224 input size).}
 \label{table:static}
 \centering
\begin{tabular}{l|cc|cccc}
  \toprule
\multirow{2}{*}{Model} & \multirow{2}{*}{FLOPs} & \multirow{2}{*}{\#Params.} & \multicolumn{4}{c}{Top-1 err.}             \\  
                       &                        &                            & ReLU & \funcA (Swish) & \funcB & \funcC      \\ \midrule
ShuffleNetV2 0.5x      & 41M                    & 1.4M                       & 39.4 & 38.3          & 38.0   & $\mathbf{37.0_{(+2.4)}}$ \\ 
ShuffleNetV2 1.5x      & 299M                   & 3.5M                       & 27.4 & 26.8          & 26.8   & $\mathbf{26.5_{(+0.9)}}$ \\ \midrule
ResNet 18              & 1.8G                   & 11.7M                      & 30.3 & 30.3          & 29.4   & $\mathbf{29.1_{(+1.2)}}$       \\ 
ResNet 50              & 3.9G                   & 25.5M                      & 24.0 & 23.5          & 23.3   & $\mathbf{23.2_{(+0.8)}}$ \\ 
ResNet 101             & 7.6G                   & 44.4M                      & 22.8 & 22.7          & 22.3   & $\mathbf{21.8_{(+1.0)}}$ \\ 
ResNet 152             & 11.3G                  & 60.0M                      & 22.3 & 22.2          & 21.7   & $\mathbf{21.2_{(+1.1)}}$ \\ 
  \bottomrule
 \end{tabular}
\end{table*}

\paragraph{\funcb}
We first evaluate our \func method on the light-weight CNNs ( MobileNets \cite{howard2017mobilenets,sandler2018mobilenetv2} and ShuffleNetV2 \cite{ma2018shufflenet}) and deep CNNs (ResNets \cite{he2016deep}).
For light-weight CNNs we follow the training configure in \cite{ma2018shufflenet};
for the larger model ResNet, we use a linear decay learning rate schedule from 0.1, a weight decay of 1e-4, a batch size of 256, and 600k iterations. 
We run numerous experiments to analyze the behavior of the \func activation function, by simply changing all the activations on various network structures and various model sizes.
The baseline networks are the ReLU networks and the extra parameters in \func networks are negligible.

We have two major observations from Table \ref{table:static} and Fig. \ref{fig:curve}.
(i), \funcA, \funcB, and \funcC all improve the accuracy remarkably comparing with their $max$-$based$ functions. This shows the benefits of the differentiable and smooth conversion.
(ii), \funcC outperforms \funcA (Swish) and \funcB, benefitting from the adaptive upper/lower bounds in the \funcC's first derivatives.
(iii), Although \funcA (Swish) shows minor improvements as the models go deeper and larger (0.1\% on ResNet-101), we still obtain continuous accuracy gains from \funcC (1.0\% on ResNet-101).

\paragraph{\funcbX}
Next, we evaluate the \funcX function.
For light-weight CNNs we change all the ReLU activations to \funcX, for deep CNN (ResNet-50, ResNet-101) we change one ReLU (after the 3$\times$3 convolution) in each building block to \funcX to avoid the overfitting problem.

The results in Table \ref{table:dynamic} show that we manage to obtain a remarkably performance boost in all the network structures.
For light-weight CNNs, \funcX improves 6.7\% on MobileNetV1 0.25 and still has around 3\% accuracy gains on 300M level models.
For deeper ResNets, \funcX still shows significant improvements, which are 2.0\% and 1.8\% on ResNet-50 and ResNet-101.

To reveal the reasons, in Fig. \ref{fig:distribution} we select a layer in the last bottleneck and compare the learned $\beta$ distribution in ResNet-50.
\func shares the same $\beta$ distribution for all the different samples across the dataset, however, in \funcX different samples have distinct non-linear degrees instead of sharing the same non-linear degree in \func.
Specifically, some samples tend to have more values close to zero, which means for such samples the network tends to have a lower non-linear degree.
While some samples tend to have more values far from zero, meaning the network adaptively learns a higher non-linearity for such samples.
This is an intuitively reasonable result as different samples usually have quite different characteristics and properties. 

\subsection{Ablation Study}
We run several ablations to analyze the proposed \func and \funcX activations.

\paragraph{Comparison with other activations}
Table \ref{table:mish} show the comparison with more activations besides ReLU and Swish, including Mish \cite{Misra2019MishAS}, ELU \cite{clevert2015fast}, SoftPlus \cite{Dugas2000IncorporatingSF}.
We note that recent advances show comparable results comparing with ReLU, except that Swish shows greater improvement (1.1\% top-1 error rate).
\func and \funcX manage to improve the accuracy remarkably (2.4\% and 4.6\%) comparing with the previous activations.

\begin{table}
 \caption{\textbf{Comparison with other activations.} We report the ImageNet top-1 error on ShuffleNetV2 0.5x.}
 \label{table:mish}
 \centering
\begin{tabular}{lc}
  \toprule
Activation & Top-1 err. \\ \midrule
ReLU  & 39.4     \\ 
Swish \cite{ramachandran2017searching}  & 38.3     \\ 
Mish \cite{Misra2019MishAS}     & 39.5      \\ 
ELU \cite{clevert2015fast}   & 39.5       \\ 
SoftPlus\cite{Dugas2000IncorporatingSF}   & 39.6       \\ \midrule
\funcC   & 37.0       \\ 
\funcX   & \textbf{34.8}       \\ 
  \bottomrule
 \end{tabular}
\end{table}
\begin{table}
 \caption{\textbf{Design space in \funcX.} We report the top-1 error on the ImageNet dataset. Comparison on 3 different levels of design space. We give 3 most simple examples on ShuffleNetV2 0.5$\times$. $fc$ denotes fully-connected, $\sigma$ denotes sigmoid, $GAP$ denotes global average pooling.}
 \label{table:ds}
 \centering
 \begin{tabular}{lcc}
  \toprule
& manner & Top-1 err.  \\ \midrule
baseline & - & 39.4 \\
pixel-wise  & $\sigma(x)$ &    37.2   \\ 
channel-wise  & $\sigma(fc[fc[GAP(x)]])$ &   34.8    \\ 
layer-wise  & $\sigma [\sum_{c}{GAP(x)}]$ &   36.3    \\ 
  \bottomrule
 \end{tabular}

\end{table}

\paragraph{Design space in \funcX}
We provide a new architecture design space in \funcX ($G(x)$ in Sec. \ref{section:meta}).
As the switching factor determines the non-linearity in the activation, we generate $\beta$ values for each sample on different levels, which could be pixel-wise, channel-wise, and layer-wise.
Our goal is to present a wide design space that provides more possibilities in the future neural network design, we are not aiming to propose the most effective specific module in this paper, which is worth more future studies.
We investigate the most simple module for each level that is described in Section \ref{section:meta}.
Table \ref{table:ds} shows the comparison on ShuffleNetV2 0.5$\times$.
The results show that all three levels could improve accuracy significantly, with more careful design, there could be more effective modules.

\paragraph{Switching factor distribution}
In \funcX we adopt a meta-learning module to learn the switching factor $\beta$ explicitly.
Figure \ref{fig:distribution} shows the distribution of the learned factor in the last activation layer of ResNet-50, we compare \funcX with \func, and randomly select 7 samples to show the result.
The distributions indicate three conclusions: (1) \funcX learns a more widespread distribution than \func; (2) each sample has its own switching factor instead of sharing the same one; (3) some samples have more values close to zero, meaning some neurons tend not to activate in this layer.

\paragraph{Comparison with SENet}
We have shown that \func helps improving accuracy remarkably. 
Next, we compare our channel-wise \funcX with the effective module SENet \cite{hu2018squeeze} on various structures.
We conduct a comprehensive comparison of both light-weight CNNs and deep CNNs.
Table \ref{table:ablate_se} shows that \funcX outperforms SENet significantly on all the network structures.
We note that it is more difficult to improve accuracy on larger networks because of highly optimized, but we find that even in the extreme deep ResNet-152, \funcX still improves accuracy by 1.8\%, which gains 1\% comparing with SENet.

Moreover, we conduct experiments on the highly optimized and extremely large network SENet-154 \cite{hu2018squeeze}, which is challenging to further improve the accuracy. 
We re-implement SENet-154 and change the activations to \func under the same experimental environment for fairness comparison.
We note that SE together with \funcA or \funcC is a case of channel-wise \funcX structure, the differences between them are the learnable upper/lower bounds (see Sec.3).
Table \ref{table:se154} shows two results: first, simply combing \funcA (Swish) with SENet performs comparable or even worse result comparing to ReLU activation; second, \funcC achieves 18.40 top-1 error rate on the challenging ImageNet dataset, improving the performance remarkably.

\begin{table}
  \caption{\textbf{\funcbX v.s. SENet} \cite{hu2018squeeze}. We report the ImageNet top-1 error rates of ShuffleNetV2 and ResNet. For fair comparison, we only replace one ReLU with \funcbX in each building block.}
 \label{table:ablate_se}
 \centering
 \begin{tabular}{lcccc}
  \toprule
& Baseline & SE & \funcX \\ \midrule
ShuffleNetV2 0.5x  & 39.4 & 37.5 & \textbf{34.8}      \\ 
ShuffleNetV2 1.5x  & 27.4 & 26.4 & \textbf{24.7}      \\ 
ResNet-50     & 24.0  & 22.8   & \textbf{22.0}     \\ 
ResNet-101   & 22.8 & 21.9 & \textbf{21.0}        \\ 
ResNet-152   & 22.3 & 21.5 & \textbf{20.5}        \\ 
  \bottomrule
 \end{tabular}
 \end{table}
 \begin{table}
  \caption{\textbf{Comparison on the extremely deep SENet-154} \cite{hu2018squeeze}. We report the ImageNet top-1 error rates. We implement all the models by ourselves.}
 \label{table:se154}
  \centering
 \begin{tabular}{lc}
  \toprule
 Activation & Top-1 err. \\ \midrule
ReLU & 18.95      \\ 
\funcA (Swish)  &  19.02    \\ 
\funcC     &    \textbf{18.40} \\ 
  \bottomrule
 \end{tabular}

\end{table}

\paragraph{More complicated activation}
In the previous sections we show the \{ \funcA, \funcB, \funcC\} activations converted from the general Maxout family.
Recently, a more powerful activation FReLU shows its potential for vision tasks.
FReLU also belongs to the Maxout family. Next, we evaluate the \funcF by simply modifying $\eta_a(x)$ and $\eta_b(x)$ according to the form of FReLU.
FReLU boosts accuracy for light-weight networks.
However, since both the FReLU layer and the original blocks contain depth-wise convolutions, directly changing the ReLU functions to FReLU is not optimal because of the redundancy utilization of the depth-wise convolution.
Therefore, to evaluate the performance, we design and train a simple toy funnel network (\fnet) made only by $point\-wise\ convolution$ and \funcF operators.
The simple block is shown in Appendix Fig. \ref{fig:tfnet}, the Appendix Table \ref{tbl:tfarch} shows the whole network structure.
We train the models with a cosine learning rate schedule, the other settings follow the work in \cite{ma2018shufflenet}.



Table \ref{table:fnet} shows the comparison to the state-of-the-art static light-weight networks.
Although the structure is very simple, the {\fnet} shows great improvements.
Since this structure does not have dynamic modules such as the SE \cite{hu2018squeeze} module, we category the {\fnet} to static networks according to the WeightNet \cite{ma2020weightnet}. 
By carefully adding dynamic modules to the structure we could get an optimal dynamic network \cite{howard2019searching}, which is beyond the focus of this work.

\begin{table}[t]
 \caption{Evaluation of \funcF, which outperforms existing static networks.}
 \label{table:fnet}
 \centering
 \resizebox{.45\textwidth}{!}{
  \begin{tabular}{lcccc}
  \toprule
   & FLOPs& \# Params. & Top-1 err. \\ \midrule
0.17 MobileNetV2  & 42M &1.4M & 52.6 \\
 ShuffleNetV2 0.5  &41M &1.4M & 39.4 \\ 
 {\fnet} 0.5 &43M &1.3M & \textbf{36.6} \\ \midrule

 0.6 MobileNetV2  & 141M  & 2.2M & 33.3  \\
 ShuffleNetV2 1.0  &146M &2.3M & 30.6 \\ 
 {\fnet}  1.0 &135M &1.9M & \textbf{29.7} \\ \midrule
 
1.0  MobileNetV2  & 300M & 3.4M & 28.0 \\
 ShuffleNetV2 1.5  &299M &3.5M & 27.4 \\ 
 {\fnet } 1.5 &279M &2.7M & \textbf{26.0} \\ \midrule

 1.4 MobileNetV2  & 585M & 5.5M & 25.3 \\
 ShuffleNetV2  2.0 &591M &7.4M & 25.0 \\ 
 {\fnet}  2.0 &\textbf{474M} &\textbf{3.8M} & \textbf{24.3} \\

  \bottomrule
 \end{tabular}
 }
\end{table}

\subsection{Generalization}
Our novel activation can easily be extended to other tasks, we show its generalization performance by experiments on object detection and semantic segmentation.

\paragraph{COCO object detection}
We report the standard COCO \cite{lin2014microsoft} metrics including AP (averaged over IoU thresholds), $AP_{50}$, $AP_{75}$, $AP_S$, $AP_M$ , $AP_L$ (AP at different scales). 
We train using the union of 80k train images and a 35k subset of validation images ($trainval35k$) and report results on the remaining 5k validation images ($minival$).
We choose the RetinaNet \cite{lin2017focal} as the detector and use ResNet-50 as the backbone. 
As a common practice, we use a batch size of 2, a weight decay of 1e-4, and a momentum of 0.9. We use anchors for 3 scales and 3 aspect ratios and use a 600-pixel train and test image scale. 
To evaluate the results of different activations, we use the ImageNet pre-trained ResNet-50 with different activations as backbones.
Table \ref{table:det} shows the significant improvements comparing with other activations.

\begin{table}[t]
 \caption{\textbf{Comparison of different activations on the COCO object detection \cite{lin2014microsoft} task}. We report results on RetinaNet \cite{lin2017focal} with ResNet-50 backbones.}
 \label{table:det}
 \centering
 \resizebox{0.45\textwidth}{!}{
  \begin{tabular}{lcccccc}
  \toprule
  & mAP & $AP_{50}$ & $AP_{75}$ & $AP_S$ & $AP_M$ & $AP_L$ \\ \midrule
ReLU    & 35.2 & 53.7 & 37.5 & 18.8 & 39.7 & 48.8  \\
Swish    & 35.8 & 54.1 & 38.7 & 18.6 & 40.0 & 49.4  \\ 
\funcX    & \textbf{36.5} & \textbf{55.9} & \textbf{38.9} & \textbf{19.9} & \textbf{40.7} & \textbf{50.6}  \\ 
  \bottomrule
 \end{tabular}
  }
\end{table}

\begin{table}[t]
 \caption{\textbf{Comparison of different activations on the CityScape \cite{cordts2016cityscapes} semantic segmentation task.} We report results on PSPNet \cite{zhao2017pyramid} with ResNet-50 backbones.}
 \label{table:seg}
 \centering
  \begin{tabular}{lcccc}
  \toprule
  Activation & FLOPs& \# Params. & mean\_IU \\ \midrule
 ReLU  & 3.9G &25.5M & 77.2 \\
 Swish   &3.9G &25.5M & 77.5 \\ 
 \funcX  &3.9G &25.7M & \textbf{78.3} \\ 
  \bottomrule
 \end{tabular}
\end{table}

\paragraph{Semantic segmentation}
We further present the semantic segmentation results on the CityScape dataset \cite{cordts2016cityscapes}.
We use the PSPNet \cite{zhao2017pyramid} as the segmentation framework and ResNet-50 as the backbone.
As a common practice, we use the poly learning rate policy where the base is 0.01 and the power is 0.9, a weight decay of 1e-4, and a batch size of 16.
Table \ref{table:seg} shows that our result (78.3) is 1.1 points higher than the ReLU baseline, showing larger improvement than Swish.
Given the effectiveness of our method on various tasks, we expect it to be a robust and effective activation for other tasks.

\section{Conclusion}
In this work, we present the novel \func as a simple but effective activation that learns to activate or not. We show \func family that approximates to the general Maxout family, exploring more functions in the Maxout family is a promising future direction yet beyond the focus of this work. 
We expect this robust and effective activation applied to a wide range of applications.


\section*{Appendix}
\renewcommand{\figurename}{Appendix Figure}
\renewcommand{\tablename}{Appendix Table}
\setcounter{figure}{0}
\setcounter{table}{0}

\begin{figure}[h]
\centering
{\includegraphics[width=.9\linewidth]{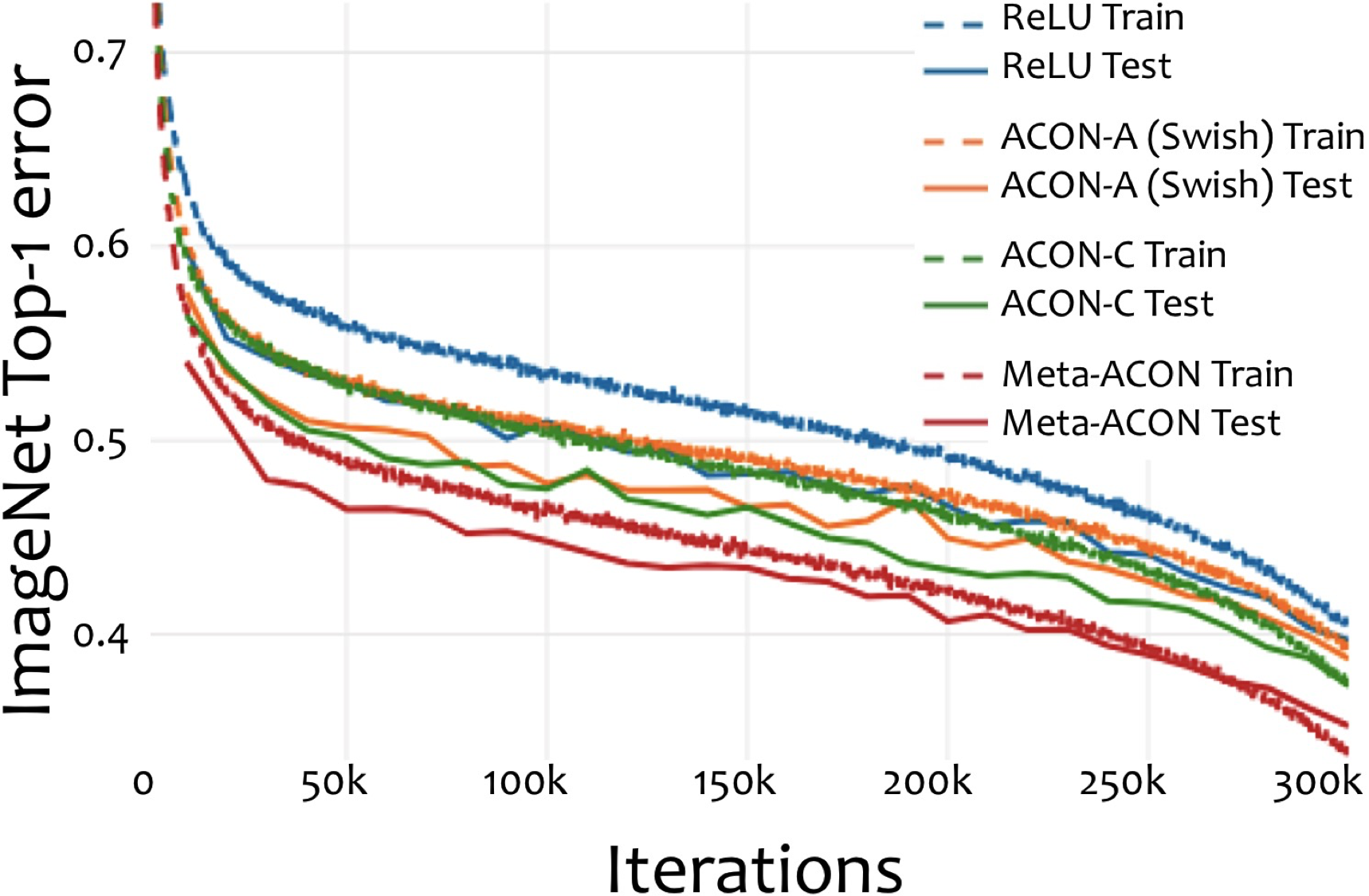}}
{\caption{The training curves. We show the ShuffleNetV2 results on the ImageNet.}
\label{fig:losscurve}}
\end{figure}

\begin{figure}[h]
\centering
\includegraphics[width=.4\textwidth,center]{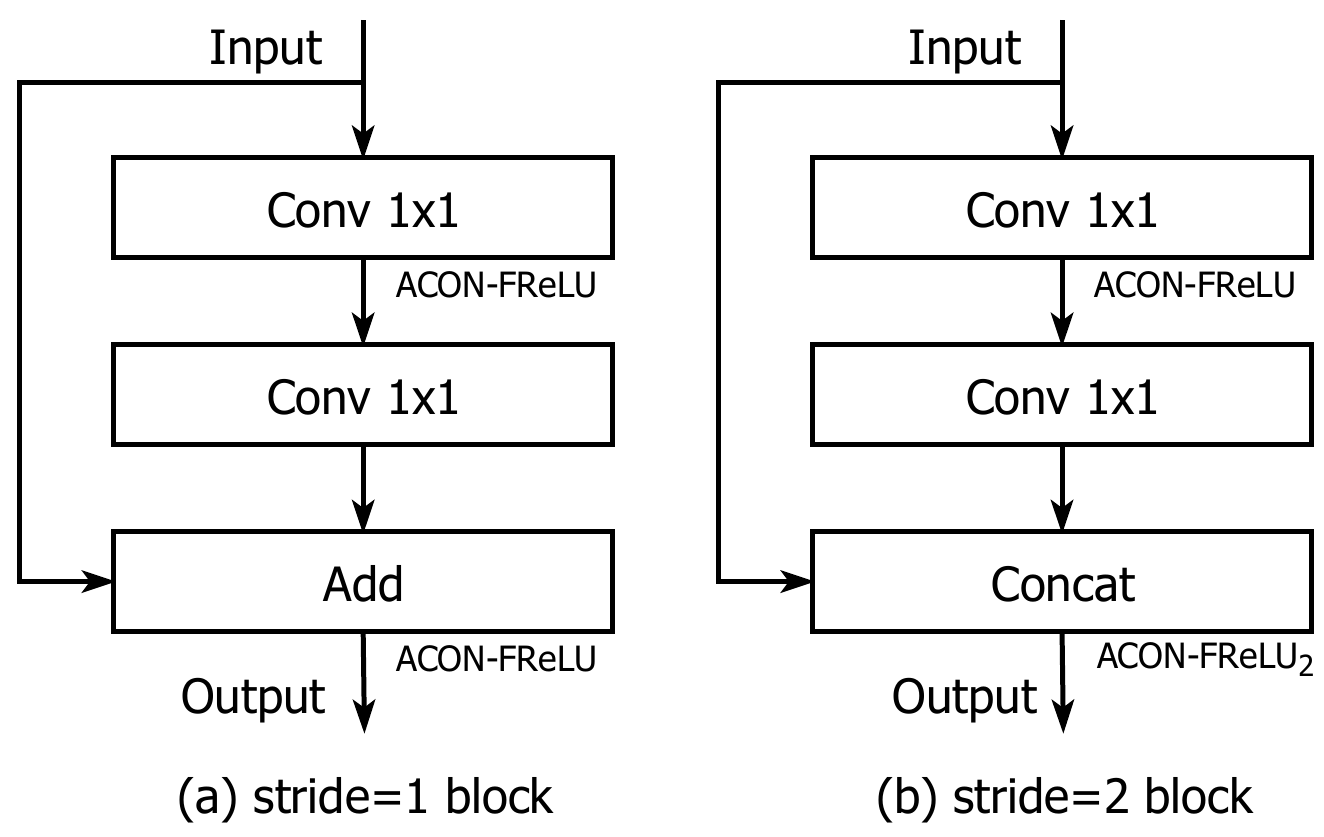}
 \caption{The simple TFNet building block. Left: stride=1, Right: stride=2. We conduct the down sampling on $\funcF_2$, which means we set $\eta_a(x)$ to be a max pooling and $\eta_b(x)$ to be a depth-wise convolution with a stride of 2.}
 \label{fig:tfnet}
\end{figure}

\begin{table}[h]
\centering
\resizebox{.5\textwidth}{!}{
\begin{tabular}{l|c|c|c|c|c|c|c|c}
\hline
\multirow{2}{*}{Layer} & \multirow{2}{*}{Output size}& \multirow{2}{*}{KSize} & \multirow{2}{*}{Stride} & \multirow{2}{*}{Repeat} & \multicolumn{4}{c}{Output channels} \\ 
\cline{6-9} & & & & & 0.5$\times$    & 1$\times$     & 1.5$\times$     & 2$\times$     \\ 
\hline
Image & 224$\times$224 & & & & 3 & 3 & 3 & 3 \\ \hline
Conv1 & 112$\times$112 & 3$\times$3 & 2 & 1 & 8 & 16 & 24 & 32 \\ \hline
Stage2 & 56$\times$56 & & \begin{tabular}[c]{@{}c@{}}2\\ 1\end{tabular} & \begin{tabular}[c]{@{}c@{}}1\\ 
1\end{tabular} & 16      & 32    & 48      & 64    \\ 
\hline
Stage3 & 28$\times$28   & & \begin{tabular}[c]{@{}c@{}}2\\ 1\end{tabular} & \begin{tabular}[c]{@{}c@{}}1\\ 3\end{tabular} & 32 & 64 & 96      & 128 \\
\hline
Stage4 & 14$\times$14 & & \begin{tabular}[c]{@{}c@{}}2\\ 1\end{tabular} & \begin{tabular}[c]{@{}c@{}}1\\ 7\end{tabular} & 64 & 128 & 192 & 256   \\ 
\hline
Stage5 & 7$\times$7 & & \begin{tabular}[c]{@{}c@{}}2\\ 1\end{tabular} & \begin{tabular}[c]{@{}c@{}}1\\ 2\end{tabular} & 128 & 256 & 384 & 512   \\ 
\hline
Conv6 & 7$\times$7 & 1$\times$1 & 1& 1 & 1024    & 1024   & 1024     & 1024   \\ 
\hline
GlobalPool & 1$\times$1 & 7$\times$7 & & & & & & \\ 
\hline
FC & & & & & 1000 & 1000   & 1000 & 1000   \\ 
\hline
FLOPs & & & & &   41M      &  146M      & 279M   & 591M \\ 
\hline
\# of Weights & & & & &   1.4M      &  2.3M      & 2.7M   & 7.4M \\ 
\hline
\end{tabular}
}
\caption{Overall architecture of the \fnet \ (toy funnel network).}
\label{tbl:tfarch}
\end{table}

{\small
\bibliographystyle{ieee_fullname}
\bibliography{egbib}
}

\end{document}